\documentclass[runningheads]{llncs}
\usepackage[T1]{fontenc}
\usepackage{graphicx,verbatim}
\usepackage{amsmath}
\usepackage[T1]{fontenc}
\usepackage{graphicx}
\usepackage{diagbox}
\usepackage{amsfonts}
\usepackage{multirow}
\usepackage{enumitem}
\usepackage{hyperref}
\usepackage[capitalise]{cleveref}

\usepackage{xcolor,soul}
\newcommand{\thickhline}{\noalign{\hrule height 1pt}}
\usepackage{float}
\usepackage{color}%, colortbl}
\usepackage{amssymb}  % For symbols like \checkmark
\usepackage{pifont}   % For \ding{55} (cross symbol)
\begin{document}
\title{Attention-based Generative Latent Replay:\newline A Continual Learning Approach for WSI Analysis}
\titlerunning{AGLR-CL for WSI Analysis}

\begin{comment}  %% Removed for anonymized MICCAI 2025 submission

\author{First Author\inst{1}\orcidID{0000-1111-2222-3333} \and
Second Author\inst{2,3}\orcidID{1111-2222-3333-4444} \and
Third Author\inst{3}\orcidID{2222--3333-4444-5555}}
\institute{Princeton University, Princeton NJ 08544, USA \and
Springer Heidelberg, Tiergartenstr. 17, 69121 Heidelberg, Germany
\email{lncs@springer.com}\\
\url{http://www.springer.com/gp/computer-science/lncs} \and
ABC Institute, Rupert-Karls-University Heidelberg, Heidelberg, Germany\\
\email{\{abc,lncs\}@uni-heidelberg.de}}

\end{comment}
\author{Pratibha Kumari\inst{1}\textsuperscript{$\star$}\textsuperscript{$@$}\and%index{Kumari, Pratibha}
Daniel Reisenb\"uchler\inst{1}\textsuperscript{$\star$}\and %index{Reisenbüchler, Daniel}
Afshin Bozorgpour\inst{1}\and %index{Bozorgpour, Afshin}
Nadine S. Schaadt\inst{2} \and %index{Schaadt, Nadine S.}
Friedrich Feuerhake\inst{2} \and %index{Feuerhake, Friedrich}
Dorit Merhof\inst{1,3} %index{Merhof, Dorit}
}
\authorrunning{Kumari et al.}
\institute{Faculty of Informatics and Data Science, University of Regensburg, Regensburg, Germany 
\and
Institute of Pathology, Hannover Medical School, Hannover, Germany 
\and 
Fraunhofer Institute for Digital Medicine MEVIS, Bremen, Germany\\
\textsuperscript{$\star$}Equal contribution, 
\textsuperscript{$@$}Correspondence (Pratibha.Kumari@ur.de)
} 
    
\maketitle              % typeset the header of the contribution
\begin{abstract}
Whole slide image (WSI) classification has emerged as a powerful tool in computational pathology, but remains constrained by domain shifts, e.g., due to different organs, diseases, or institution-specific variations.
To address this challenge, we propose an \textbf{A}ttention-based \textbf{G}enerative \textbf{L}atent \textbf{R}eplay \textbf{C}ontinual \textbf{L}earning framework (AGLR-CL), in a multiple instance learning (MIL) setup for \textit{domain incremental} WSI classification. Our method employs Gaussian Mixture Models (GMMs) to synthesize WSI representations and patch count distributions, preserving knowledge of past domains without explicitly storing original data. A novel attention-based filtering step focuses on the most salient patch embeddings, ensuring high-quality synthetic samples. This privacy-aware strategy obviates the need for replay buffers and outperforms other buffer-free counterparts while matching the performance of buffer-based solutions. We validate AGLR-CL on clinically relevant biomarker detection and molecular status prediction across multiple public datasets with diverse centers, organs, and patient cohorts. Experimental results confirm its ability to retain prior knowledge and adapt to new domains, offering an effective, privacy-preserving avenue for domain incremental continual learning in WSI classification.

% Authors must provide keywords and are not allowed to remove this Keyword section.

\end{abstract}

\section{Introduction}
%old
% The rapid advancements in computational pathology and artificial intelligence (AI) have led to significant progress in computer-aided diagnosis for histopathology image analysis, enabling automated disease detection and biomarker assessment. Whole Slide Image (WSI) classification is a particularly challenging task due to the high resolution of WSIs and the lack of localized annotations. To address this, multiple instance learning (MIL) has emerged as a widely adopted weakly supervised learning paradigm, effectively aggregating patch-level features into slide-level predictions using only slide-level annotations.
%
Recent advances in computational pathology (CPath) and digitizing WSIs have transformed histopathology image analysis, driving significant progress in automated disease detection and biomarker assessment. However, WSI classification remains challenging due to the gigapixel resolution and the lack of pixel-level annotations. A common strategy divides WSIs into manageable patches, which are processed offline by vision encoding models to obtain feature sequences. Notably, self-supervised pretraining has enabled the development of domain-specific foundation models (FMs) that outperform out-of-domain counterparts~\cite{xu2024gigapath,chen2024towards,wang2023retccl}, such as ImageNet-pretrained models. The conversion of patch-level features into slide-level predictions is achieved through MIL by aggregation of these features.

%old
% Despite these advancements, WSI classification models face several challenges in real-world clinical applications. Variability in histopathological features across organ-specific biology, staining protocols and scanner manufacturers, and study cohorts introduces distribution shifts that can significantly degrade model performance when applied to new, unseen datasets. Conventional MIL models are typically trained on a fixed dataset and deployed under the assumption that the data distribution remains unchanged. However, in practice, domain shifts often arise due to differences in WSIs obtained from distinct hospitals, patient populations, or acquisition settings, leading to poor generalization of static models. One approach to mitigate distribution shifts is fine-tuning the MIL model on new data, but this often results in catastrophic forgetting, where the model loses knowledge of previously learned distributions. Retraining the model from scratch with both past and new data is another alternative, yet it is impractical due to the massive storage demands of WSIs (which are gigapixel-sized) and the escalating computational costs as data accumulates over time.
%

\noindent Despite these advancements, WSI classification models still face challenges in clinical settings. Variability in morphological features, originating from differences in organ-specific biology, staining protocols, scanner manufacturers, and patient cohorts, induces distribution shifts that degrade performance on new datasets. Conventional MIL models struggle to generalize when WSIs are acquired from diverse hospitals and settings. Fine-tuning on new datasets is a common adaptation strategy; however, it often leads to catastrophic forgetting (CF)~\cite{kirkpatrick2017overcoming,KUMARI2024117100}. On the other hand, continual learning (CL) has emerged as a promising solution for evolving medical data while mitigating CF~\cite{kumari2023continual,Kum_Continual_MICCAI2024}. By enabling continuous knowledge accumulation, CL enhances model robustness and adaptability in clinical settings and facilitates forward knowledge transfer, e.g., from frequently stained datasets in H\&E or PAS to those for follow-up diagnostics like CD8 or TRI~\cite{Kum_Continual_MICCAI2024}. Although buffer-based methods, which store selected past samples, typically yield superior performance~\cite{derakhshani2022lifelonger,bhatt2024characterizing}, their applicability to WSIs is hindered by storage and privacy constraints. Existing WSI CL research is limited, focusing primarily on buffer-based and class incremental methods~\cite{huang2023conslide,zhu2024lifelong}.
\noindent To address these limitations, we propose AGLR-CL, a buffer-free generative replay approach for domain incremental WSI classification. AGLR-CL models past domain distributions with GMMs trained on patch embeddings and counts. For each domain, class-wise multivariate GMMs and one-dimensional GMMs capture prior data distribution. In subsequent domains, synthetic data sampled from these GMMs are combined with new WSIs to update the MIL model, thus avoiding real data storage and preserving privacy. We validate AGLR-CL on multiple tasks across domain incremental datastes including various centers and organs. Extensive experiments show that AGLR-CL effectively retains prior knowledge and adapts to new domains, surpassing other buffer-free methods and achieving performance close to buffer-based methods. 
%%%without storing real data, thus increasing privacy. 
Our main contributions are: 

\noindent \textbf{(1) Domain incremental CL for MIL.} To our knowledge, we introduce domain incremental CL for MIL for the first time and present a GMM and attention-based filtering for effective re-sampling of past data across domains. 

\noindent \textbf{(2) Broad applicability and increased privacy.} Across CPath tasks, including biomarker screening and molecular status predictions, our AGLR-CL consistently surpasses buffer-free methods and is on par with buffer-based methods, while avoiding WSI storage and thus increasing privacy.

% \begin{itemize}[leftmargin=*,label={}]
% \item \textbf{(1) Domain incremental CL for MIL.} To our knowledge, we introduce domain incremental CL for MIL for the first time and present a GMM and attention-based filtering for effective re-sampling of past data across domains.
% %\vspace{0.5em}
% % \item \textbf{(2) Broad applicability and increased privacy.} Across CPath tasks, including biomarker screening and molecular status predictions, our AGLR-CL framework consistently surpasses buffer-free methods and is on par with buffer-based methods, while avoiding sample storage and thus increasing privacy. 
% \item \textbf{(2) Broad applicability and increased privacy.} Across CPath tasks, our AGLR-CL consistently surpasses buffer-free methods and is on par with buffer-based methods, while avoiding WSI storage and thus increasing privacy. 
% \end{itemize}

% \begin{enumerate}[left=0pt]
% \item \textbf{Domain incremental CL for MIL.} To our knowledge, we introduce domain incremental CL for MIL for the first time and present a GMM and attention-based filtering for effective re-sampling of past data across domains.
% \item \textbf{Broad applicability and increased privacy.} Across CPath tasks our AGLR-CL framework consistently surpasses buffer-free methods and is on par with buffer-based methods without storing samples and thus increasing privacy. 
% \end{enumerate}

%%%%%%%%%%%%%%%%%%%%%%%%%%%%%%%%%%%%%%%%%%%%%%%%%%%%%%%%%%%%%%%%%%
\begin{figure*}[!ht]
\centering
\includegraphics[scale=0.6]{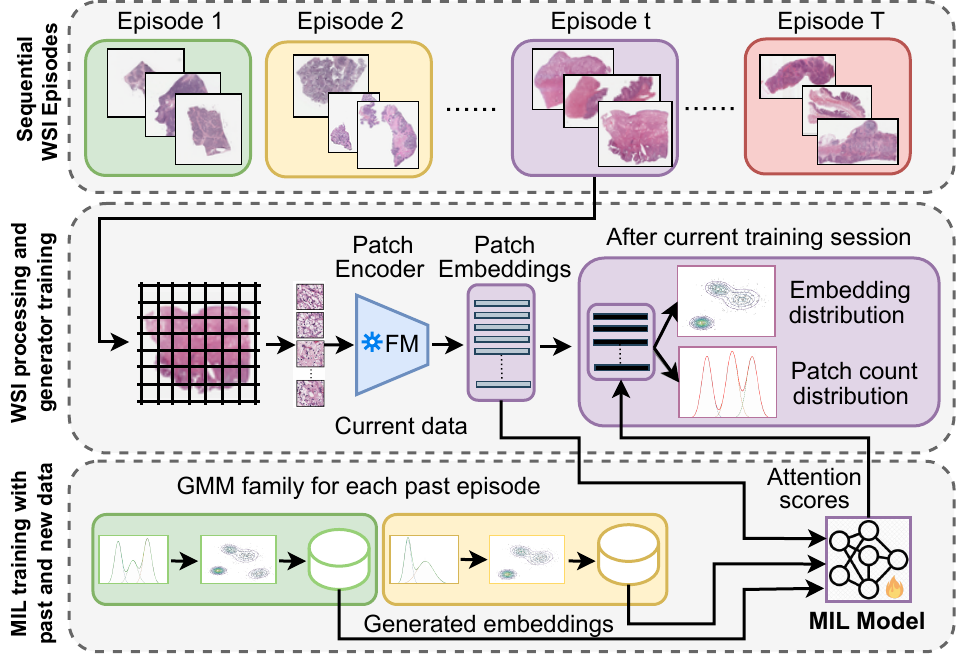}
\caption{\textbf{Privacy-aware domain incremental AGLR-CL framework.} Following WSI tessellation, a frozen FM-based encoder generates a sequence of tractable embeddings. A MIL aggregator is then trained on the current episode, with high-attention features selected to fit GMMs for patch embeddings and counts. In subsequent episodes, historical data is revisited by re-sampling synthetic WSIs using the per-episode GMMs.}
\label{fig:flowchart}
\end{figure*}
%%%%%%%%%%%%%%%%%%%%%%%%%%%%%%%%%%%%%%%%%%%%%%%%%%%%%%%%%%%%%%%%%%%%
\section{Method}
% A flowchart of the proposed approach is shown in Fig.~\ref{fig:flowchart}. In the following we detail MIL-based WSI Classification, continual learning settings and our AGLR-CL framework. This strategy consists of GMM-based synthetic embedding generation that leverages attention and the generative latent replay mechanism.
A flowchart of the proposed approach is shown in Fig.~\ref{fig:flowchart}. 
In the following, we detail MIL-based WSI classification, CL settings, and our AGLR-CL framework, which incorporates an attention-based selection for GMM training and synthetic embedding generation for a latent replay mechanism.

\subsection{MIL-based WSI Classification}
We adopt a standard preprocessing pipeline, 
%%where each WSI is partitioned into $n$ non-overlapping patches 
partitioning each WSI into $n$ non-overlapping patches $p_i \in \mathbb{R}^{512\times 512\times 3}$. A pretrained CPath FM is then used to extract patch embeddings, resulting in a feature sequence $\{f_i\}_{i=1}^{n} \in \mathbb{R}^{n \times D}$, with $D$ denoting the latent dimension. These embeddings are aggregated using a learnable MIL model $\mathcal{M}$. Specifically, we employ AB-MIL~\cite{ilse2018attention}, which embeds each feature into a lower-dimensional space $d$ via a linear layer and applies an attention mechanism to assign instance-level weights. The weighted embeddings are summed and fed into a classification head for WSI prediction.

%old
% \subsection{Continual Learning Settings}
% In continual WSI classification applications, WSI datasets $(D_1, D_2, \dots D_T)$ are to be learned sequentially by the model $\mathcal{M}$. These sequence datasets, also called episodes, potentially represent different domains. At any given time $t$, the training set of only the current dataset $D_t$ is available while the test set of all the episodes ($D_1, D_2, \dots D_t, \dots D_T$) is always available. 
% Although some works~\cite{zhu2024lifelong,huang2023conslide} allow storing a few past samples in a buffer memory to update the model along with the new data, we aim to deal with catastrophic forgetting in a buffer-free manner, which becomes important in privacy-constrained situations. 
% To achieve this, for each encountered episode, we train class-wise multivariate GMMs on the collected patch embeddings and one-dimensional (1D) GMMs on the patch counts across all the WSIs in the dataset. Then, these generators can be used to mimic past domains, on demand by genrating latent vectors. The classification model $M$ is updated with current as well as generated synthetic data to recap past as well as learn the new domain. Below, we first discuss the WSI classification pipeline, then describe the GMM-based synthetic WSI embedding generation, and finally, the generative replay mechanism that mitigates forgetting.

\subsection{Continual Learning Configuration}
We consider a CL pipeline for WSI classification, where datasets arrive sequentially (defined as episodes), $\{\mathcal{D}_1, \mathcal{D}_2, \ldots, \mathcal{D}_T\}$, each representing a distinct domain $t \in \{1, ..., T\}$. At training domain $t$, the model $\mathcal{M}$ has access to the current dataset $\mathcal{D}_t$ only, while evaluation is performed on test sets from all episodes $\{\mathcal{D}_1, \mathcal{D}_2, \ldots, \mathcal{D}_T\}$. Unlike approaches that retain a buffer of past samples~\cite{zhu2024lifelong,huang2023conslide}, our method addresses CF while avoiding WSI storage, a critical requirement in privacy-sensitive domains. 

\subsection{GMM-based Synthetic Embedding Generation}\label{sec:generator}
WSIs inherently contain a variable number of patches. To generate synthetic WSI representations, we model both the patch counts and patch embeddings using GMMs. For a new dataset $\mathcal{D}_t$, we estimate class-specific multivariate models $\text{GMM}^{t}_{\text{emb}}$ on the collected patch embeddings to capture feature variations. Concurrently, one-dimensional models $\text{GMM}^{t}_{\text{count}}$ are fitted on patch counts to account for tissue variability across WSIs. Since not all patches contribute meaningfully to the classification task, we introduce an attention-guided filtering step prior to GMM estimation. After the $t^{\text{th}}$ training session with classifier $\mathcal{M}_{t}$, attention scores are computed for patches across each WSI in the dataset of the current episode $\mathcal{D}_t$. Low-attention patches are discarded, retaining only the top $q\%$ for subsequent processing. Consequently, the feature sequence $\{f_i\}_{i=1}^{n_j}$ of each WSI $\mathcal{W}_j \in \mathcal{D}_t$ with $j\in \{1,\dots, \lvert \mathcal{D}_t \rvert\}$ is reduced to $m_j<n_j$. 
Next, we define GMMs for WSI embedding generation. The probability density function for a feature embedding is its likelihood under a $K$-component GMM, given by
%

 %\begin{align*}
 \begin{equation}
 \begin{aligned}
p(f_i | \Theta) = \sum_{k=1}^{K} \pi_k \mathcal{N}(f_i | \mu_k, \Sigma_k),
\end{aligned}
\end{equation}
%\end{align*}

%
%
\noindent where $\mathcal{N}(f_i | \mu_k, \Sigma_k)$ represents a Gaussian density function with $k$ mixture parameters given by mean $\mu_k$ and covariance $\Sigma_k$, which are defined by
 \begin{equation}
 \begin{aligned}
\mu_k = \frac{\sum_{i=1}^{n} \gamma_{ik} f_i}{\sum_{i=1}^{n} \gamma_{ik}}, \quad \Sigma_k = \frac{\sum_{i=1}^{n} \gamma_{ik} (f_i - \mu_k)(f_i - \mu_k)^T}{\sum_{i=1}^{n} \gamma_{ik}},
 \end{aligned}
 \end{equation}

\noindent with the responsibility $\gamma_{ik}$ computed via the Expectation-Maximization~\cite{dempster1977maximum} to update the parameters $\mu_k, \Sigma_k$ iteratively:
%%%%%We use the Expectation-Maximization (EM) algorithm~\cite{dempster1977maximum} to update the parameters $\mu_k, \Sigma_k$ iteratively,
%
 \begin{equation}
 \begin{aligned}
\gamma_{ik} = \frac{\pi_k \mathcal{N}(f_i | \mu_k, \Sigma_k)}{\sum_{j=1}^{K} \pi_j \mathcal{N}(f_i | \mu_j, \Sigma_j)}.
 \end{aligned}
 \end{equation}

\noindent The optimal $K$ is selected by minimizing the Bayesian Information Criterion (BIC)~\cite{fraley1998many} over candidate values. 
%%%%%%The suitable value of $K$ is determined using the Bayesian Information Criterion (BIC)~\cite{fraley1998many}. BIC values are tracked for a set of candidate $K$ values and the one that minimizes BIC is selected. 
The estimated parameters define a generative model that facilitates on-the-fly sampling of synthetic patch embeddings mimicking $\mathcal{D}_{t}$.
%%%%, thereby reinforcing past knowledge while learning new episodes. 
Concurrently, the number of patches $\hat{n}_j$ in a synthetic WSI $\mathcal{W}_j \in \mathcal{D}_{t}$ is determined by sampling from $\text{GMM}^{t}_{\text{count}}$, ensuring that the generated WSIs have a realistic patch count, or in other words, tissue variability. We denote the union of GMMs created for each episode as GMM family.

% \subsection{Generative Replay}
% During any $t^{th}$ training session, the training dataset $D_t$ is available along with the trained GMMs from all previous sessions. To mitigate catastrophic forgetting, we generate synthetic WSIs representing past domains using these GMMs. Specifically, for each past session $t' < t$, to generate a WSI embedding of shape $N'\times C$, we sample a patch count $N'$ from $\text{GMM}_{t'}^{\text{count}}$ and generate $N'$ embeddings from $\text{GMM}_{t'}^{\text{emb}}$. The count of total synthetic WSIs are same the current dataset size while maintaining the class ratio observed in $D_{t'}$. In total, WSI embeddings for past session ${t'}$ are generated in the same amount as the current session dataset but following the class ration, as in $D_{t'}$.
% These generated WSIs are then combined with the real WSIs from the current dataset. The hybrid dataset, which includes both freshly acquired and synthetically replayed samples, is used to train the AttentionMIL classifier. 
% By integrating synthetic data through generative replay, the model continuously reinforces knowledge from previous episodes. This mitigates catastrophic forgetting, prevents overfitting to newly acquired data, and eliminates the need to store real historical samples, making the approach highly suitable for privacy-sensitive applications.

\subsection{Generative Latent Replay}
During the \(t^{\text{th}}\) training session, the current dataset \(\mathcal{D}_t\) is expanded with synthetic WSIs generated from the $t-1$ GMM families learned in all previous episodic datasets $\{\mathcal{D}_1, \mathcal{D}_2, \ldots, \mathcal{D}_{t-1}\}$. For each past session \(t' < t\), synthetic WSI embeddings are generated as feature sequences $\{f_i\}_{i=1}^{\hat{n}_j}$ for $j\in \{1,\dots, \lvert \mathcal{D}_{t'} \rvert\}$. To this end, we first sample a patch count $\hat{n}_j$ from $\text{GMM}^{t'}_{\text{count}}$ and subsequentially drawing $\hat{n}_j$ patch embeddings from $\text{GMM}^{t'}_{\text{emb}}$. The number of synthetic WSIs matches the WSIs count in \(\mathcal{D}_t\) while preserving the class ratio previously observed in \(\mathcal{D}_{t'}\). These synthetic samples are then combined with the real WSIs from the current session to form a hybrid training set. By integrating synthetic data through generative replay, the model continuously reinforces knowledge from previous episodes, thereby mitigating CF, reducing overfitting to new data, and eliminating the need to store real historical samples.
\begin{table*}[!htbp]
\centering
\caption{\textbf{Dataset statistics.} Overview of data cohorts across organs, centers, and tasks, to create both homogeneous and heterogeneous domain shifts. We used a patient-stratified split into Train/Test sets to avoid data leakage from individual datasets.}
\label{tab:allDatasets}
\begin{tabular}{@{} c@{} c | >{\centering\arraybackslash}c@{\hspace{10pt}} >{\centering\arraybackslash}c@{\hspace{10pt}} | c c @{}}
\thickhline
 & \textbf{Name} & \begin{tabular}[c]{@{}c@{}}\textbf{Train}\\Class 0/1\end{tabular} & \begin{tabular}[c]{@{}c@{}}\textbf{Test}\\Class 0/1\end{tabular} & \textbf{Organ} & \textbf{Center} \\ \hline

\parbox[t]{3mm}{\multirow{5}{*}{\rotatebox[origin=c]{90}{\textbf{MSI}}}} 
 &  TCGA-CRC~\cite{TCGA} & 303/52 & 79/13 & Colorectal & multiple \\
 &  CPTAC-COAD~\cite{CPTAC} & 138/41 & 30/12 & Colon & $C1$ \\
 &  PAIP-CRC~\cite{PAIP} & 28/9 & 7/3 & Colorectal & $C2$ \\
 &  TCGA-STAD~\cite{TCGA} & 239/48 & 62/12 & Stomach & multiple \\
 & TCGA-UCEC~\cite{TCGA} & 340/92 & 88/25 & Uterine & multiple \\ \hline
 
\parbox[t]{3mm}{\multirow{5}{*}{\rotatebox[origin=c]{90}{\textbf{TMB}}}} 
 &  TCGA-STAD~\cite{TCGA} & 261/66 & 67/16 & Stomach & multiple \\
 &  TCGA-UCEC~\cite{TCGA} & 278/152 & 68/38 & Uterine & multiple \\
 & TCGA-NSCLC~\cite{TCGA} & 533/280 & 143/67 & Lung & multiple \\
  &  TCGA-CRC~\cite{TCGA} & 350/65 & 90/16 & Colorectal & multiple \\
 & TCGA-BRCA~\cite{TCGA} & 853/26 & 210/6 & Breast & multiple \\ \hline
\parbox[t]{3mm}{\multirow{3}{*}{\rotatebox[origin=c]{90}{\textbf{HER2}}}} 
 & TCGA-BRCA~\cite{TCGA} & 469/135 & 114/32 & Breast & multiple \\
 &  CPTAC-BRCA~\cite{CPTAC} & 266/38 & 56/7 & Breast & $C4$ \\
 &  BCNB~\cite{BCNB} & 625/221 & 156/56 & Breast & $C5$ \\ \hline
\parbox[t]{3mm}{\multirow{3}{*}{\rotatebox[origin=c]{90}{\textbf{PR}}}} 
 &  TCGA-BRCA~\cite{TCGA} & 275/577 & 71/145 & Breast & multiple \\
 &  CPTAC-BRCA~\cite{CPTAC} & 114/159 & 41/34 & Breast & $C4$ \\
 & BCNB~\cite{BCNB} & 214/632 & 54/158 & Breast & $C5$ \\ 
\thickhline
\end{tabular}
\end{table*}
%%%%%%%%%%%%%%%%%%%%%%%%%%%%%

%%%%%%%%%%%%%%%%%%%%%%%%%%%%%%%%%%%%%%%%%%%%%%%%%%%%%%%%%%%%%%%%
\section{Experiments}\label{sec:exp}
\noindent \textbf{Datasets.} We consider multiple publicly available WSI datasets for biomarker screening of microsatellite instability (MSI) and tumor mutational burden (TMB), 
%%%%%where we obtained binarized labels by a threshold for the numerical TMB values of $10$ mutations/megabase. 
binarizing TMB numeric values at 10 mutations/megabase.
We also perform molecular status prediction of progesterone receptor (PR) and human epidermal growth factor receptor 2 (HER2) in breast cancer. Specifically, we explore data repositories such as The Cancer Genome Atlas (TCGA)~\cite{TCGA}, Clinical Proteomic Tumor Analysis Consortium (CPTAC)~\cite{CPTAC}, PAIP2020 challenge~\cite{PAIP}, and Early Breast Cancer Core-Needle Biopsy WSI (BCNB)~\cite{BCNB}. For TCGA and CPTAC, we obtain labels from \href{https://cbioportal.org}{cbioportal.org}. 
Table~\ref{tab:allDatasets} summarizes the datasets, detailing tasks, train/test volume, organs, and centers. Datasets from multiple centers are marked \textit{multiple}; otherwise, labeled as $(C1, C2, \dots)$.
%%% Table~\ref{tab:allDatasets} presents all the datasets with details about training and testing volume, classification tasks, organ, and data curation center. If a dataset involves multiple data centers, then it is marked \textit{multiple}, otherwise we refer to each distinct data center as $(C1, C2, \dots)$.

\noindent\textbf{Continual Learning Episodes.} To comprehensively evaluate AGLR-CL, we create multiple sequences from datasets listed in Table~\ref{tab:allDatasets}, each having multiple datasets as episodes. WSI datasets in each sequence exhibit differences in terms of organ, center, and mixed shifts to create domain incremental scenarios in the CL framework. The sequences are detailed in Table~\ref{tab:DATASETsEQUENCE}.

%%%%%%%%%%%%%%%%%%%%%%%%%%%%%%%%%%%%%%%%%%%%%%%%%%%%%%%%%%%%%
\begin{table*}[!ht]
\centering
% \caption{\textbf{Dataset sequences as episodes for domain shifts.} We apply heterogeneous (a2, a1, a5) and center (a3, a4) shifts in data to obtain domain incremental episodes. %Further sequence permutations and evaluations are provided in the Supplementary Material.
% }
\caption{\textbf{Dataset episodes detail.} We curate heterogeneous organ/center (a2, a1, a5) and homogeneous center (a3, a4) shifts to obtain domain incremental episodes. }

\label{tab:DATASETsEQUENCE}
\resizebox{\textwidth}{!}{%
\begin{tabular}{@{} c l l @{}}
\thickhline
\textbf{Seq.} & \textbf{Task} & \textbf{Dataset episodes} \\
\hline
a1 & MSI &  TCGA-STAD $\rightarrow$ PAIP-CRC $\rightarrow$ TCGA-UCEC \\
a2 & MSI &  PAIP-CRC $\rightarrow$ TCGA-CRC $\rightarrow$ TCGA-STAD $\rightarrow$ CPTAC-COAD $\rightarrow$   TCGA-UCEC \\
% a2-old & MSI & \scriptsize TCGA-CRC $\rightarrow$ CPTAC-COAD $\rightarrow$ TCGA-STAD $\rightarrow$ PAIP-CRC $\rightarrow$ TCGA-UCEC \\
a3 & PR &  TCGA-BRCA $\rightarrow$ CPTAC-BRCA $\rightarrow$ BCNB \\
a4 & HER2 &  TCGA-BRCA $\rightarrow$ BCNB $\rightarrow$ CPTAC-BRCA \\
a5 & TMB &  TCGA-STAD $\rightarrow$  TCGA-UCEC $\rightarrow$  TCGA-NSCLC $\rightarrow$ TCGA-CRC $\rightarrow$  TCGA-BRCA \\
\thickhline
\end{tabular}%
}
\end{table*}
%===============================================================

%%%%%%%%%%%%%%%%%%%%%%%%%%%%%%%%%%%%%%%%%%%%%%%%
\begin{table*}[!htbp]
\centering
\fontsize{8pt}{10pt}\selectfont
\caption{\textbf{Performance comparison across CL methods.} Past raw data (PRD) marks if past WSIs need to be stored. \textcolor{red}{Red} and \textcolor{blue}{blue} indicate first and second best performances by all CL methods. \underline{Underline} shows best performance in buffer‐free CL.}
\label{tab:comp_cl_all}
\begin{tabular}{c|c|c|c|ccc|ccc|ccc}
\thickhline
\multirow[c]{2}{*}{\bf Task} & \multirow[c]{2}{*}{\bf Seq.} & \multirow[c]{2}{*}{\bf PRD} & \multirow[c]{2}{*}{\bf Method} & \multicolumn{3}{c|}{\textbf{weighted F1}} & \multicolumn{3}{c|}{\textbf{AUROC}} & \multicolumn{3}{c}{\textbf{AUPRC}} \\
& & & & ACC & ILM & BWT & ACC & ILM & BWT &  ACC & ILM & BWT \\\hline

\multirow{8}{*}{MSI}   & \multirow{8}{*}{a1} & \ding{55} & Naive & 65.20 & 65.45 & -17.32 & 61.17 & 66.7  & -23.67 & 32.31 & 39.93 & -25.97 \\
&                    & \ding{51}   & Joint   & 69.84 & —   & —    & 65.96 & —   & —    & 41.21 & —   & — \\
& & \ding{51} & Cumulative & 76.02 & 76.71 & -0.41 & 77.63 & 81.6  & -6.13  & 54.85 & 59.23 & -7.67 \\
&  & \ding{51} & Replay~\cite{rolnick2019experience} & \textcolor{red}{73.02} & \textcolor{red}{75.07} & \textcolor{blue}{-2.84} & \textcolor{red}{67.87} & \textcolor{red}{73.22} & \textcolor{blue}{-5.27} & \textcolor{red}{47.88} & \textcolor{blue}{50.44} & \textcolor{red}{-5.51} \\
&   & \ding{51} & GDumb~\cite{prabhu2020gdumb} & 64.52 &65.34 &\textcolor{red}{-0.5}
&58.54 &61.91 &\textcolor{red}{-0.19}&34.96 &36.83 &\textcolor{blue}{-5.55}\\
&                    & \ding{55} & LwF~\cite{radio3} & 59.94 & 64.53 & -11.72 & 53.42 & 59.95 & -31.47 & 31.68 & 40.22 & \underline{-23.5} \\
&                    & \ding{55} & EWC~\cite{kirkpatrick2017overcoming} & 64.17 & 70.06 & -21.79 & 55.16 & 65.74 & -33.83 & 37.27 & 46.57 & -27.65 \\
&                    & \ding{55} & SI~\cite{zenke2017continual} & 66.25 & 68.15 & -16.64 & 60.26 & 65.77 & -37.24 & \textcolor{blue}{\underline{45.24}} & \textcolor{red}{\underline{51.74}} & -23.68 \\
&                    & \ding{55} & \textbf{Proposed} & \textcolor{blue}{\underline{69.91}} & \textcolor{blue}{\underline{73.84}} & \underline{-4.45} & \textcolor{blue}{\underline{64.88}} & \textcolor{blue}{\underline{72.87}} & \underline{-20.18} & 38.38 & 49.44 & -23.84 \\\hline

\multirow{8}{*}{MSI}   & \multirow{8}{*}{a2} & \ding{55} & Naive &68.19 &74.59 &-2.63&70.82 &76.33 &1.27&45.44 &54.46 &6.78 \\ 
&    & \ding{51}      & Joint &79.95 &- &-&74.07 &- &-&56.87 &- &-  \\
&   & \ding{51} & Cumulative&72.90 &76.51 &3.36&72.13 &76.53 &4.03&41.11 &50.8 &2.92 \\
&    & \ding{51} & Replay~\cite{rolnick2019experience} &68.34 &72.35 &-2.59&\textcolor{blue}{68.32} &\textcolor{red}{75.01} &\textcolor{blue}{-1.6}&\textcolor{blue}{40.51} &47.24 &-3.41 \\
&   & \ding{51} & GDumb~\cite{prabhu2020gdumb}&\textcolor{blue}{71.66} &68.05 &\textcolor{red}{16.11}&60.69 &60.42 &\textcolor{red}{3.39}&40.47 &38.56 &\textcolor{blue}{2.19}  \\
&    & \ding{55} & LwF~\cite{radio3} &68.71 &\textcolor{blue}{72.68} &-3.86&62.51 &70.76 &-5.0&34.76 &42.47 &-6.39 \\ 
&   & \ding{55} & EWC~\cite{kirkpatrick2017overcoming}&65.03 &70.49 &-7.01&59.17 &69.81 &-9.16&37.52 &45.9 &-1.26 \\
&   & \ding{55} & SI~\cite{zenke2017continual} &68.58 &71.53 &-7.1&63.51 &70.68 &-8.5&36.97 &\textcolor{blue}{47.26} &-5.66 \\
&   & \ding{55} & \textbf{Proposed} &\textcolor{red}{\underline{78.01}} &\textcolor{red}{\underline{74.1}} &\textcolor{blue}{\underline{-1.89}}&\textcolor{red}{\underline{69.11}} &\textcolor{blue}{\underline{74.05 }}&-2.32&\textcolor{red}{\underline{47.66}} &\textcolor{red}{\underline{52.94}} &\textcolor{red}{\underline{9.61}}\\\hline

% \multirow{8}{*}{MSI}   & \multirow{8}{*}{a2-old} & \ding{55} & Naive & 68.66 & 73.08 & -8.41 & 65.78 & 70.85 & -12.02 & 38.35 & 46.61 & -13.18 \\ 
% &    & \ding{51}      & Joint   & 80.25 & —   & —    & 78.58 & —   & —    & 60.90 & —   & — \\
% &   & \ding{51} & Cumulative & 78.24 & 79.7  & -0.73 & 73.14 & 74.85 & 3.4    & 54.49 & 55.01 & 5.35 \\
% &    & \ding{51} & Replay~\cite{rolnick2019experience} & \textcolor{blue}{75.46} & \textcolor{red}{79.64} & \textcolor{blue}{-3.88} & \textcolor{blue}{68.33} & \textcolor{red}{75.87} & -6.89 & \textcolor{blue}{48.97} & \textcolor{red}{55.63} & \textcolor{blue}{-4.23} \\
% &   & \ding{51} & GDumb~\cite{prabhu2020gdumb} & 69.19 &71.5 & \textcolor{red}{5.47}&63.12 &60.42 & \textcolor{red}{9.42}&38.54 &34.09 & \textcolor{red}{5.29} \\
% &                    & \ding{55} & LwF~\cite{radio3} & 66.42 & 75.95 & -11.69 & 61.42 & 71.7  & \textcolor{blue}{\underline{-6.03}} & 38.14 & 48.24 & -10.36 \\ 
% &                    & \ding{55} & EWC~\cite{kirkpatrick2017overcoming} & 64.63 & 74.75 & -12.44 & 67.26 & \textcolor{blue}{\underline{75.35}} & -10.57 & 37.69 & 51.31 & -13.13 \\
% &                    & \ding{55} & SI~\cite{zenke2017continual} & 59.77 & 70.4  & -12.97 & 47.85 & 62.61 & -17.77 & 26.90 & 38.9  & -17.84 \\
% &                    & \ding{55} & \textbf{Proposed} & \textcolor{red}{\underline{76.46}} & \textcolor{blue}{\underline{78.99}} &\underline{-7.53} & \textcolor{red}{\underline{69.00}} & 73.9  & -10.01 & \textcolor{red}{\underline{49.91}} & \textcolor{blue}{\underline{52.69}} & \underline{-9.26} \\\hline

\multirow{8}{*}{PR}    & \multirow{8}{*}{a3} & \ding{55} & Naive & 64.29 & 67.29 & -7.44 & 66.85 & 69.08 & -9.85 & 73.34 & 74.06 & -8.04 \\
&                    & \ding{51}      & Joint   & 69.76 & —   & —    & 71.84 & —   & —    & 77.69 & —   & — \\
&                    & \ding{51} & Cumulative & 67.60 & 66.73 & 0.37 & 71.40 & 72.0  & -0.5   & 77.73 & 77.31 & -0.48 \\
&                    & \ding{51} & Replay~\cite{rolnick2019experience} & \textcolor{red}{70.77} & \textcolor{red}{67.76} & \textcolor{blue}{3.54} & 68.48 & \textcolor{blue}{69.18} & -4.56 & 74.96 & 74.31 & -2.46 \\
&   & \ding{51} & GDumb~\cite{prabhu2020gdumb} &67.21 &63.22 & \textcolor{red}{11.52}& \textcolor{red}{72.82} &67.97 & \textcolor{red}{9.7}& \textcolor{red}{77.51} &\textcolor{blue}{74.98} &\textcolor{red}{2.5}\\
&    & \ding{55} & LwF~\cite{radio3} & 66.91 & 64.1  & -2.05 & \textcolor{blue}{\underline{70.35}} & 68.16 & \textcolor{blue}{\underline{-1.79}} & 76.00 & 73.28 & \textcolor{blue}{\underline{-0.65}} \\
&   & \ding{55} & EWC~\cite{kirkpatrick2017overcoming} & 63.28 & 64.27 & -4.82 & 66.30 & 66.87 & -7.13 & 73.36 & 72.13 & -5.03 \\
&    & \ding{55} & SI~\cite{zenke2017continual} & 63.55 & 65.03 & -6.61 & 66.77 & 68.28 & -6.64 & 74.72 & 74.34 & -3.42 \\
&                    & \ding{55} & \textbf{Proposed} & \textcolor{blue}{\underline{67.97}} & \textcolor{blue}{\underline{66.33}} & \underline{-1.48} & 70.34 & \textcolor{red}{\underline{71.16}} & -8.0 & \textcolor{blue}{\underline{76.35}} & \textcolor{red}{\underline{76.23}} & -6.28 \\\hline

\multirow{8}{*}{HER2}   & \multirow{8}{*}{a4} & \ding{55} & Naive & 71.80 & 71.99 & -3.21 & 59.57 & 63.6  & -5.41  & 29.36 & 35.53 & -5.98 \\
&                    & \ding{51}       & Joint   & 75.85 & —   & —    & 62.90 & —   & —    & 36.44 & —   & — \\
&                    & \ding{51} & Cumulative & 75.59 & 75.66 & 0.84 & 61.82 & 66.07 & 1.96 & 36.91 & 41.5  & 5.1 \\
&  & \ding{51} & Replay~\cite{rolnick2019experience} & 75.32 & \textcolor{blue}{74.12} & \textcolor{blue}{1.7} & 59.24 & 63.93 & \textcolor{blue}{1.59} & 31.95 & \textcolor{red}{38.0} & \textcolor{blue}{1.62} \\
&   & \ding{51} & GDumb~\cite{prabhu2020gdumb} &\textcolor{blue}{75.64} &73.43 &\textcolor{red}{3.56}&\textcolor{blue}{62.13} &64.77 &\textcolor{red}{8.47}&\textcolor{blue}{32.95} &37.69 &\textcolor{red}{7.5}\\
&                    & \ding{55} & LwF~\cite{radio3} & 72.29 & 72.16 & -4.59 & 58.12 & 62.71 & -5.04 & 28.07 & 34.21 & -5.44 \\
&   & \ding{55} & EWC~\cite{kirkpatrick2017overcoming} & 71.44 & 72.31 & -5.39 & 61.22 & \textcolor{blue}{64.97} & -2.85 & 28.90 & 35.74 & -5.34 \\
&                    & \ding{55} & SI~\cite{zenke2017continual} & 71.62 & 72.74 & -3.39 & 55.37 & 62.21 & -0.53 & 28.27 & 35.92 & \underline{-1.58} \\
&                    & \ding{55} & \textbf{Proposed} & \textcolor{red}{\underline{76.94}} & \textcolor{red}{\underline{75.11}} & \underline{-0.78} & \textcolor{red}{\underline{66.18}} & \textcolor{red}{\underline{66.79}} & \underline{-0.49} & \textcolor{red}{\underline{35.07}} & \textcolor{blue}{\underline{37.99}} & -1.97 \\\hline

\multirow{8}{*}{TMB}   & \multirow{8}{*}{a5} & \ding{55} & Naive & 70.78 & 68.84 & -7.16 & 49.50 & 59.27 & -11.33 & 22.84 & 35.24 & -7.17 \\
&                    &\ding{51}      & Joint   & 74.35 & —   & —    & 61.48 & —   & —    & 34.49 & —   & — \\
&                    & \ding{51} & Cumulative & 74.48 & 69.75 & 0.51 & 55.40 & 60.4  & -0.03 & 32.91 & 37.88 & 1.95 \\
&                    & \ding{51} & Replay~\cite{rolnick2019experience} & 71.62 & \textcolor{blue}{69.4} & \textcolor{blue}{-2.09} & 50.77 & 60.39 & -7.39 & 26.53 & 36.84 & \textcolor{blue}{-4.44} \\
&   & \ding{51} & GDumb~\cite{prabhu2020gdumb} &\textcolor{blue}{72.98} &67.89 &\textcolor{red}{0.49}&54.72 &55.47 &\textcolor{red}{4.76}&28.08 &33.37 &\textcolor{red}{3.07}\\
&                    & \ding{55} & LwF~\cite{radio3} & 70.32 & 69.05 & \underline{-4.42} & \textcolor{blue}{55.51} & \textcolor{blue}{60.58} & \textcolor{blue}{\underline{-6.12}} & \textcolor{blue}{28.35} & \textcolor{blue}{37.93} & \underline{-5.11} \\
&                    & \ding{55} & EWC~\cite{kirkpatrick2017overcoming} & 70.63 & 67.32 & -11.63 & 48.34 & 59.13 & -15.18 & 20.60 & 33.09 & -13.94 \\
&                    & \ding{55} & SI~\cite{zenke2017continual} & 70.44 & 67.76 & -5.29 & 43.98 & 56.4  & -13.12 & 20.33 & 34.53 & -7.99 \\
&                    & \ding{55} & \textbf{Proposed} & \textcolor{red}{\underline{73.04}} & \textcolor{red}{\underline{69.58}} & -4.73 & \textcolor{red}{\underline{57.45}} & \textcolor{red}{\underline{62.39}} & -10.6 & \textcolor{red}{\underline{31.53}} & \textcolor{red}{\underline{38.53}} & -5.25 \\
\thickhline
\end{tabular}%
\end{table*}
%%%%%%%%%%%%%%%%%%%%%%%%%%%%%%%%%%%%%%%%%%%%%%%%%%%%%%

\noindent\textbf{Continual Learning Baselines.} We compare our method against various CL baselines, including regularization methods such as EWC~\cite{kirkpatrick2017overcoming}, SI~\cite{zenke2017continual}, and LwF~\cite{radio3} and rehearsal methods such as GDumb~\cite{prabhu2020gdumb} and Replay~\cite{rolnick2019experience}. 
%%%%We report lower bound performance by a naive approach, where the model is fine-tuned with only current episode data and upper bound performances by joint and cumulative training, where all the episode data are assumed to be available for training at the same time.
Further, we report lower bound performance by the naive approach and upper bound performances by joint and cumulative approaches. {\it Naive} corresponds to traditional fine-tuning with only current episode data, {\it joint} uses all datasets simultaneously, and {\it cumulative} sequentially incorporates all previous datasets.

\noindent\textbf{Implementation Details.}
We extracted patches using the CLAM library~\cite{lu2021data} and employed the pre-trained UNI~\cite{chen2024towards} pathology FM for feature extraction. 
The buffer for Replay and GDumb is set to $100$. 
%%%%We set a buffer size of $100$ for GDumb and Replay. 
For SI, EWC, and LwF, the regularizing factor ($\alpha$) was set to 1 by following the literature~\cite{Kum_Continual_MICCAI2024}. In our AGLR-CL, we keep $q$ as $80\%$.
%%%In our AGLR-CL, we keep the threshold for filtering out patches with low-attention patches as $q=80\%$. 
%%%%The number of components in GMM for patch embedding is searched in [8, 24] with step size of 8 and in GMM built on patch counts as [1,5] with step size of 1. 
%%%%The suitable number of components ($K$) in GMMs are learned using the Bayesian Information Criterion (BIC)~\cite{fraley1998many}. Specifically, we track BIC for GMMs with $K$ in [8, 24] and [1, 5] for $\text{GMM}^{}_{\text{emb}}$ and $\text{GMM}^{}_{\text{count}}$, respectively and the one that minimizes BIC is selected. 
We select $K$ from $\{8, 16, 24\}$ for $\text{GMM}^{t}_{\text{emb}}$ and $\{1, 2, 3, 4, 5\}$ for $\text{GMM}^{t}_{\text{count}}$.
To accommodate for class imbalances, we track weighted F1 score, AUROC, and AUPRC metrics. For sequential training and evaluation in CL with $T$ episodes, we consider $T \times T$ as train-test matrix~\cite{kumari2023continual} where cell $T_{ij}$ denote  performance on $j^{th}$ datasets after $i^{th}$ training session with $\mathcal{D}_i$. We compute CL metrics from $T \times T$ matrix, including forgetting measure BWT~\cite{diaz2018don} and average performances using ACC~\cite{lopez2017gradient}, computed at the last episode and ILM~\cite{diaz2018don,kumari2023continual}, computed at every training session. The larger these metrics, the better the performance. All experiments were conducted using a single NVIDIA H100 GPU.

%%%%%%%%%%%%%%%%%%%%%%%%%%%%%%%%%%%%%%%%%%%%%
\section{Results}\label{sec:results}

Table~\ref{tab:comp_cl_all} compares AGLR-CL against competing approaches on MSI prediction, PR status, HER2 status, and TMB mutation. We report ACC, ILM, and BWT based on weighted F1, AUROC, and AUPRC. Across sequences a1–a5, the naive update of the model exhibits lower performance (ACC and ILM) and higher forgetting (BWT) compared to the cumulative, while joint training on all data provides an upper bound. Among CL methods, buffer-based methods (Replay and GDumb) generally achieve the best performance (red), with our approach following in a1 and a3 and surpassing them in a2, a4 and a5. Notably, when considering buffer-free methods only, our approach mostly delivers the best results (\underline{underlined}). Thus, while slightly trailing buffer-based methods, our buffer-free solution offers a competitive alternative in privacy-sensitive applications.

\begin{figure}[t]
\centering
\includegraphics[width=\textwidth, keepaspectratio]{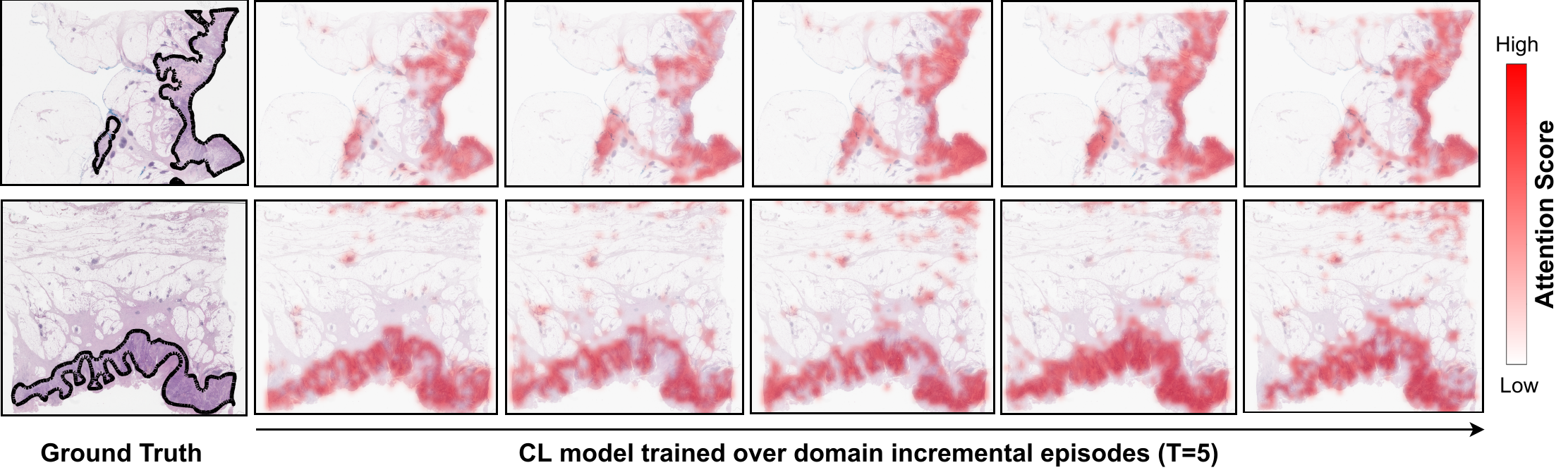}
\caption{\textbf{Attention heatmaps for AGLR-CL across domain shifts in MSI prediction.} Attentions scores for WSIs from $\mathcal{D}_1$ (PAIP-CRC) by the model trained over five CL stages in a2 reflect that past knowledge acquired from $\mathcal{D}_1$ is preserved.}
\label{fig:CL_heatmaps}
\end{figure}

\noindent 
% Fig.~\ref{fig:CL_heatmaps} shows attention heatmaps for two WSIs from $\mathcal{D}_1$ (PAIP-CRC) by model $\mathcal{M}$, following a  sequential training with $\{\mathcal{D}_1, \mathcal{D}_2, \ldots, \mathcal{D}_5\}$ in $a2$ sequence. 
Fig.~\ref{fig:CL_heatmaps} shows attention heatmaps for two WSIs from $\mathcal{D}_1$ (PAIP-CRC) in $a2$ sequence by model $\mathcal{M}$ over five training sessions, corresponding to sequential training with different datasets.
It can be observed that high attention scores cover the annotated region in all CL training sessions. Interestingly, an organ-shift $(t=3,5)$ creates a few artifacts compared to center-only shifts $(t=2,4)$.
Overall, consistent attention to ground-truth area reflects that past knowledge is preserved while learning on new datasets with differences in centers and organs.
%%%%, with a few artifacts in sessions t=3,5 due to the multitude of domain shifts (organ and center) as compared to sessions (t=2 and 4) with only center shifts. 

%===============================================================
\begin{table}[]
\centering
\caption{\textbf{Ablation study for attention-based filtering (ABF).} Best in \textbf{bold}.}
\label{tab:abalation}
\begin{tabular}{c|c|c|ccc|ccc|ccc}
\thickhline
\multirow[c]{2}{*}{\bf Task} & \multirow[c]{2}{*}{\bf Seq.} & \multirow[c]{2}{*}{\bf w/o ABF} & \multicolumn{3}{c|}{\bf weighted F1} & \multicolumn{3}{c|}{\bf AUROC} & \multicolumn{3}{c}{\bf AUPRC} \\[1ex]
&  &   & ACC & ILM & BWT & ACC & ILM & BWT &  ACC & ILM & BWT \\ \hline
                          
\multirow{2}{*}{MSI} & \multirow{2}{*}{a1} 
& \ding{55} & 67.13 & 69.15 & -13.8 & 63.09 & 68.51 & -24.22 & \textbf{41.61} & 45.44 & -24.21 \\
&    & \ding{51} & \textbf{69.91} & \textbf{73.84} & \textbf{-4.45} & \textbf{64.88} & \textbf{72.87} & \textbf{-20.18} & 38.38 & \textbf{49.44} & \textbf{-23.84} \\ \hline

\multirow{2}{*}{MSI} & \multirow{2}{*}{a2} 
& \ding{55}  &66.16 &67.75 &-9.84&68.59 &73.30 &\textbf{-1.22}&45.57 &49.27 &8.17 \\
&& \ding{51}&\textbf{78.01} &\textbf{74.10} &\textbf{-1.89}&\textbf{69.11} &\textbf{74.05} &-2.32&\textbf{47.66} &\textbf{52.94} &\textbf{9.61}\\ \hline

% \multirow{2}{*}{MSI} & \multirow{2}{*}{a2-old} 
% & \ding{55} & 68.54 & 75.18 & -9.01 & \textbf{70.58} & 73.2 & -12.31 & 46.59 & 47.9 & -11.04 \\
% % &  & \ding{51} & \textbf{75.46} & \textbf{79.64} & \textbf{-3.88} & \textbf{68.33} & \textbf{75.87} & \textbf{-6.89} & \textbf{48.97} & \textbf{55.63} & \textbf{-4.23} \\ \hline
% && \ding{51}&\textbf{76.46} &\textbf{78.99} &\textbf{-7.53}
% &69.00 &\textbf{73.9} &\textbf{-10.01}
% &\textbf{49.91} &\textbf{52.69} &\textbf{-9.26}\\ \hline

\multirow{2}{*}{PR}  & \multirow{2}{*}{a3} 
& \ding{55} & 64.32 & 63.72 & -3.14 & 62.93 & 65.00 & \textbf{-6.92} & 69.85 & 70.90 & -8.5 \\
&   & \ding{51} & \textbf{67.97} & \textbf{66.33} & \textbf{-1.48} & \textbf{70.34} & \textbf{71.16} & -8.0 & \textbf{76.35} & \textbf{76.23} & \textbf{-6.28} \\ \hline

\multirow{2}{*}{HER2} & \multirow{2}{*}{a4} 
& \ding{55} & 73.12 & 73.41 & -5.73 & 61.75 & 65.06 & -5.65 & 29.32 & 36.07 & -7.79 \\
& & \ding{51} & \textbf{76.94} & \textbf{75.11} & \textbf{-0.78} & \textbf{66.18} & \textbf{66.79} & \textbf{-0.49} & \textbf{35.07} & \textbf{37.99} & \textbf{-1.97} \\ \hline

\multirow{2}{*}{TMB} 
& \multirow{2}{*}{a5} 
& \ding{55} & \textbf{73.58} & \textbf{70.32} & \textbf{-1.88} & \textbf{58.60} & \textbf{65.39} & \textbf{-2.95} & 30.98 & \textbf{38.86} & \textbf{-0.09} \\
&   & \ding{51} & 73.04 & 69.58 & -4.73 & 57.45 & 62.39 & -10.6 & \textbf{31.53} & 38.53 & -5.25 \\ 
\thickhline
\end{tabular}%
\end{table}

%============================================================
%============================================================

%%%https://app.diagrams.net/#G1xdAguovaeMlA4c2-4vhRu-U0Eb0albh1#%7B%22pageId%22%3A%22wbQVnTjbejIUsQyL5iBJ%22%7D

% \noindent\textbf{Ablation.} Table \ref{tab:abalation} presents an ablation study on attention-based filtering for GMMs training. 
% %%%We report ACC, ILM, and BWT computed from a train-test matrix based on weighted F1 score, AUROC, and AUPRC. 
% The results show that, except for sequence a5, filtering consistently outperforms direct training on all patch embeddings. For sequence a5, the slight drop is probably due to the high variability of morphological changes caused by different TMB outcomes across organs, thus discarding some patches can hurt data recovery. Optimizing the filtering mechanism to account for organ variability may improve performance in more challenging scenarios.

\noindent\textbf{Ablation.} Table \ref{tab:abalation} presents an ablation study on attention-based filtering for GMMs training. 
The results show that, except for sequence a5, GMMs trained with filtered patches consistently outperform those trained on all patch embeddings. For sequence a5, the slight drop may occur due to the high variability of morphological alterations in different TMB outcomes across organs, thus discarding certain patches can hurt data recovery. 
%Optimizing the filtering mechanism to account for organ variability may improve performance in more challenging scenarios.

\section{Conclusion}
We proposed AGLR-CL, a buffer-free generative latent replay framework enabling privacy-aware CL for WSI tasks including biomarker screening and molecular status predictions. Instead of maintaining a buffer, AGLR-CL leverages GMMs to synthesize past feature distributions, allowing the model to retain knowledge while adapting to new domains. Results demonstrate that AGLR-CL mitigates CF and achieves state-of-the-art performance in privacy-sensitive CL.

\bibliographystyle{splncs04}




% LNCS Homepage, \url{http://www.springer.com/lncs}, last accessed 2023/10/25
% \end{thebibliography}
\bibliography{main}

\end{document}